\documentclass[preprint,12pt]{elsarticle}

\usepackage{graphicx}%
\usepackage{multirow}%
\usepackage{amsmath,amssymb,amsfonts}%
\usepackage{amsthm}%
\usepackage{mathrsfs}%
\usepackage[title]{appendix}%
\usepackage{xcolor}%
\usepackage{textcomp}%
\usepackage{manyfoot}%
\usepackage{booktabs}%
\usepackage{algorithm}%
\usepackage{algorithmicx}%
\usepackage{algpseudocode}%
\usepackage{listings}%
\usepackage{lmodern}
\usepackage{subfigure}
\usepackage{xfrac}

\journal{Information Sciences}

\begin{document}
	\begin{frontmatter}
	
		\title{A Multi-Channel Spatial-Temporal Transformer Model for Traffic Flow Forecasting}
		
		\author{Jianli Xiao\corref{cor1}}
		\ead{audyxiao@sjtu.edu.cn}
		
		\author{Baichao Long}
		
		\address{School of Optical-Electrical and Computer Engineering, University of Shanghai for Science and Technology, Shanghai, 200093, China}
		
		\cortext[cor1]{Corresponding author.}
		
		\begin{abstract}
			Traffic flow forecasting is a crucial task in transportation management and planning. The main challenges for traffic flow forecasting are that (1) as the length of prediction time increases, the accuracy of prediction will decrease; (2) the predicted results greatly rely on the extraction of temporal and spatial dependencies from the road networks. To overcome the challenges mentioned above, we propose a multi-channel spatial-temporal transformer model for traffic flow forecasting, which improves the accuracy of the prediction by fusing results from different channels of traffic data. Our approach leverages graph convolutional network to extract spatial features from each channel while using a transformer-based architecture to capture temporal dependencies across channels. We introduce an adaptive adjacency matrix to overcome limitations in feature extraction from fixed topological structures. Experimental results on six real-world datasets demonstrate that introducing a multi-channel mechanism into the temporal model enhances performance and our proposed model outperforms state-of-the-art models in terms of accuracy.
		\end{abstract}
		
		\begin{keyword}
			
			multi-channel \sep graph convolutional network \sep Transformer \sep traffic flow forecasting
			
		\end{keyword}
		
	\end{frontmatter}
	
	%% \linenumbers
	
	%% main text
	\section{Introduction}
	With the rapid development of urbanization, urban traffic has become a pressing issue that needs to be addressed. The increasing number of vehicles in cities has led to problems such as road congestion, traffic accidents, and environmental pollution becoming increasingly severe. Traffic flow forecasting, an important component of intelligent transportation systems, provides technical support for alleviating traffic congestion, reducing traffic accidents, and lowering environmental pollution \cite{b1}. However, due to the complex spatial and temporal dependencies of traffic flow, traffic flow forecasting remains a challenging problem. In recent years, most research has used deep learning to model traffic flow. Graph neural networks (GNNs) \cite{b2} and convolutional neural networks (CNNs) \cite{b3} are commonly used for modeling on the spatial scale. Recurrent neural networks (RNNs) \cite{b4} including its variants such as long short-term memory (LSTM) \cite{b5} and gated recurrent unit (GRU) \cite{b99} are commonly used for modeling on the temporal scale. With the popularity of Transformers, which have also been used for traffic flow forecasting tasks, but the complex spatial-temporal correlations of traffic flow are still difficult to learn.
	
	Traffic flow forecasting models can be divided into three types: parametric models, non-parametric models, and hybrid models. Early statistical models such as auto-regressive integrated moving average models \cite{NEW1}, grey prediction models \cite{NEW2}, and Kalman filter models \cite{NEW3} are typical parameter models, which rely on static assumptions of the system and could not reflect the nonlinearity and uncertainty of traffic data. Support vector regression models \cite{NEW4}, \emph{K}-nearest neighbor models \cite{NEW5}, Bayesian models \cite{NEW6}, and fuzzy-neural models \cite{NEW7} are classical non-parametric models. All of these models only consider the temporal dependence of traffic flow and ignore the spatial dependence between roads. This makes traffic flow forecasting results unconstrained by the road network, and it is unable to accurately predict actual traffic conditions. Hybrid models are composed of multiple models combined to solve multi-scale dependence problems. Due to the excellent ability of graph convolutional networks (GCNs) \cite{b14} to extract spatial features, they are widely utilized in traffic flow forecasting tasks. Many researchers have combined spatial models with temporal models to form hybrid models for spatial-temporal prediction tasks, which simultaneously consider both spatial and temporal dependencies \cite{b6}.
	
	\begin{figure*}
		\centerline{\includegraphics[width=1\textwidth]{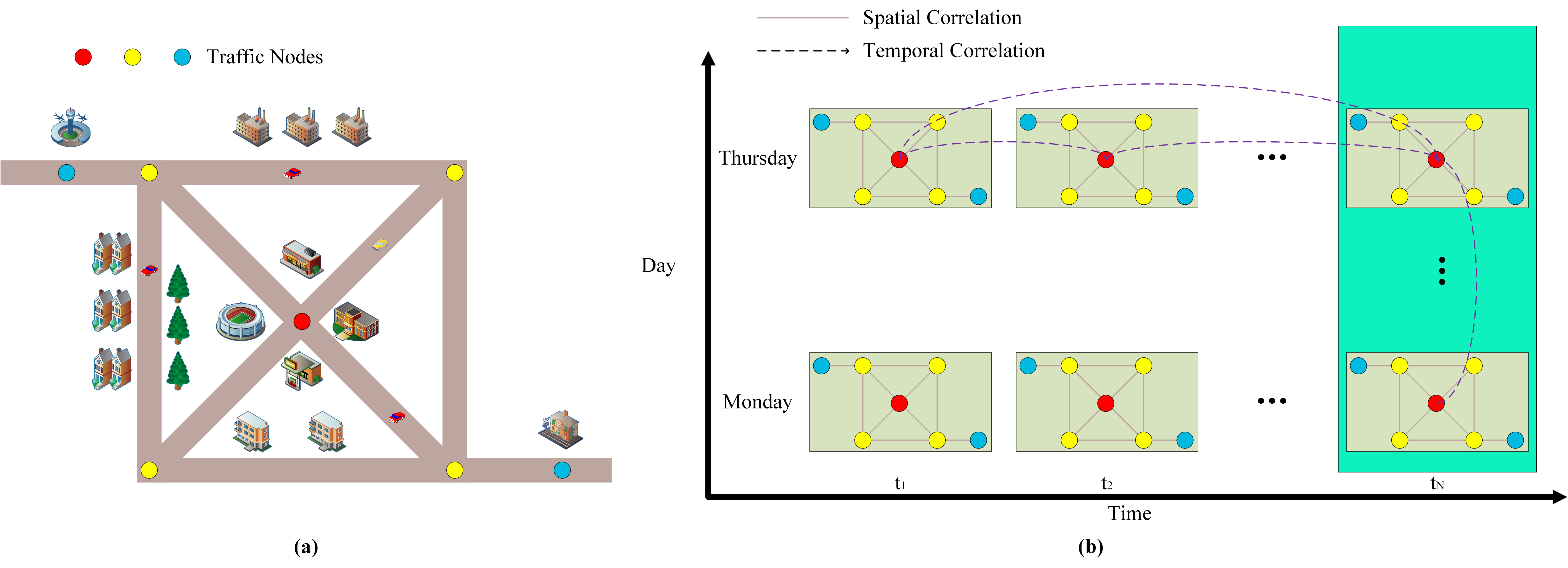}}
		\caption{Visualization of spatial-temporal correlation: \textbf{(a)} distribution of urban road and traffic sensors, where red, yellow, and blue dots represent traffic sensors at different locations with different traffic conditions; \textbf{(b)} time slices of road network from time $t_1$ to $t_N$, where the current traffic flow at each node is not only related to the traffic conditions in the previous slice of the day, but also related to the historical traffic flow at the same time on previous days.}
		\label{fig1}
	\end{figure*}
	
	The traffic flow in urban road networks exhibits randomness, uncertainty, and periodicity. This phenomenon is primarily influenced by factors such as accidents, changes in city planning, and fluctuations in population. The interaction of these factors leads to variations and fluctuations in traffic flow, which are not only related to adjacent time slices but also to previous daily, weekly, and monthly patterns at that time. As shown in Fig. \ref{fig1}(a), the red dot represents a node of school road. Due to students' fixed arrival and departure times, the traffic conditions at the same location during peak period exhibit long-term similarities. Therefore, when predicting traffic flow, it is necessary to consider not only short-term temporal correlations but also long-term temporal correlations, which means taking into account the impact of multiple time channels on the results. The arrival and departure times of other schools within the road network are generally synchronized, so even if the distance between different schools is considerable, their traffic flow will exhibit a certain correlation. Therefore, when predicting traffic flow, it is necessary to consider not only spatial correlations in terms of distance but also dynamic spatial correlations. This requires incorporating not only static graph structures but also dynamic graph structures in GCNs. As illustrated in Fig. \ref{fig1}(b), we can predict the traffic conditions at time $t_N$ based on the traffic data from time $t_1$ to $t_{N-1}$ on Thursday, taking into account the temporal dependence of consecutive time slices as mentioned earlier. Additionally, we need to incorporate data from the same time slice on Wednesday, Tuesday, and Monday to predict the traffic conditions at this time, considering the temporal dependencies across different days. To achieve the aforementioned object, we propose a new Multi-Channel Spatial-Temporal Model (MC-STTM) for traffic flow forecasting. Data are extracted from multiple channels and passed through the model for spatial-temporal feature traffic flow forecasting. For spatial-temporal feature extraction, GCNs and Transformers \cite{b15} are used. In GCNs, both the adjacency matrix generated from known topological structures and the adaptive adjacency matrix are jointly utilized to extract spatial features of road networks, capturing spatial dependencies more accurately. In the position encoding of Transformer, not only all time intervals within a day are encoded, but each day of the week from Monday to Sunday is also encoded.
	
	In summary, the distinction of our model from previous studies lies in the introduction of a multi-channel mechanism to account for the impact of cyclical factors on traffic flow forecasting. At the spatial scale, we consider both static and dynamic spatial dependencies to fully capture spatial information. At the temporal scale, although the Transformer architecture is currently very popular, there are still some studies using other deep time series models. In this paper, we compare the Transformer architecture with other deep time series models before and after the introduction of the multi-channel mechanism as a way to justify our choice of Transformer. Previous studies have focused on the combination of time series models with spatial models, and few have compared the models at different scales. Therefore, MC-STTM is proposed in this paper. We evaluate our approach on six public traffic datasets, and the experimental results demonstrate that our model achieves promising performance compared to state-of-the-art baseline models. The main contributions of this paper are as follows:
	\begin{itemize}
		\item We propose a multi-channel data fusion model for traffic flow forecasting, which models the traffic network by incorporating historical traffic data from different channels, outperforming existing traffic flow forecasting models that do not leverage multi-channel data.
		
		\item We utilize GCNs to extract spatial features, where the construction of the adjacency matrix differs from previous models. We combine an adaptive adjacency matrix with a known road network topology-generated adjacency matrix to further capture spatial dependencies.
		
		\item We utilize Transformers to extract temporal features, which exhibit competitive performance in long-term prediction compared to RNNs. We also introduce different position encoding mechanisms in different channels to further capture temporal dependencies.
		
		\item We conduct comparative experiments on multiple datasets, including both traffic flow datasets and speed datasets. The results show that our model outperforms baseline models.
	\end{itemize}
	
	\textbf{Paper Organization}. In Section 2, we provide a detailed review of the main developments in traffic flow forecasting. In Section 3, we present MC-STTM framework in detail. Subsequently, Section 4 outlines the experimental design and results. Lastly, in Section 5, we summarize our findings and conclusions.
	
	\section{Related Works}
	In this section, we present a comprehensive review of the primary research on traffic flow forecasting. The existing models for predicting traffic flow can be broadly classified into three categories: temporal models that rely on temporal-based scales, spatial models that rely on spatial-based scales, and hybrid models that combine both temporal and spatial scales.
	\subsection{Temporal Models}
	With the rapid development of deep learning, traditional models are not commonly used in existing traffic flow forecasting tasks. RNNs and their variants, such as LSTM and GRU, have become commonly used time series models in recent years. 
	Transformers are the most popular deep learning network architecture in recent years, mainly used for sequence-to-sequence learning tasks. In time series prediction, we can use Transformers to learn the corresponding mapping relationships. Compared with RNNs, Transformers achieves good results in handling long-term dependencies and can better capture relationships between sequences through position encoding and other models. Transformers can also be computed in parallel and can handle input sequences of any length, making the model more flexible. In summary, as the development of time series models has progressed, Transformers have become the most advantageous time series network model architecture for traffic flow forecasting tasks.
	\subsection{Spatial Models}
	Time series prediction models only consider temporal dependencies, neglecting spatial dependencies, which results in inaccurate traffic flow forecasting that do not take into account the constraints imposed by road networks. Some scholars have introduced CNNs into spatial dependency models, which are highly effective in processing regular grid data. However, road network structures are often irregular graph structures, and CNNs, which are inherently suited to Euclidean spaces, have limitations when dealing with the complex topologies of traffic road networks. They are fundamentally unable to describe spatial dependencies. Compared to CNNs, GCNs perform better in processing non-Euclidean structured data, as they can capture the structural features of graph data. As the two models face different data structures, GCNs are widely used for traffic flow forecasting on road networks and achieve good results. For graph models, we need to know sufficient information about the topology of road network graphs, such as adjacency matrices and feature matrices, meaning that adequate topological information about road networks is needed to adopt spatial models. To solve the problem of incomplete topological information for spatial feature extraction, some scholars have proposed models such as adaptive adjacency matrices. In summary, the model of GCNs has become the most popular model for traffic flow forecasting tasks.
	\subsection{Hybrid Models}
	In the current research field, hybrid methods combining time series and spatial dependency models have emerged as powerful methods for analyzing and forecasting traffic flow. This approach is particularly suited for extracting temporal characteristics and spatial structural information from road network data. In recent years, researchers have developed advanced models by integrating time series and spatial dependency models, achieving significant progress in this field.
	Spatio-Temporal Graph Convolutional Network (STGCN) \cite{b16} is a typical model among these, employing a graph-based approach instead of traditional convolutional and recurrent structures to handle the non-linearity and complexity of traffic flow. However, as Lu and Li \cite{b17} note, traditional GNNs have limitations in modeling the time-varying spatial-temporal correlations among sensors and often rely on predefined graph structures, which are not always available or entirely reliable. To overcome this challenge, Topologically Enhanced Spatial-Temporal Graph Convolutional Network \cite{b19} was designed to capture spatial dependencies in data more effectively. Another approach, Trajectory WaveNet \cite{b18}, addresses the shortcomings of graph-based methods in capturing complex dependencies between continuous road segments, such as prohibited left turns and dynamic spatial dependencies.
	Liu et al. \cite{b33} combines spatial-temporal similarity feature modules with spatial-temporal convolution modules that incorporate attention mechanisms. This combination effectively solves the challenge of extracting spatial-temporal relations from complex traffic flows. Zhang et al. \cite{b24} emphasize that simple and fixed spatial graph structures, relying only on prior knowledge of the traffic network, can lead to poor predictive performance. Similarly, Zheng and Zhang \cite{b20} point out that static graphs cannot directly learn dynamic spatial-temporal dependencies across time periods. Yang et al. \cite{b23} adopted a novel method combining spatial-temporal convolution and attention mechanisms to learn extensive spatial-temporal correlations. The proposed local spacetime neural network method aims to capture local traffic patterns without relying on a specific network structure. Multi-Scale Temporal Dual Graph Convolution Network \cite{b21} combining gated spatial-temporal convolution and dual graph convolution modules, effectively extracting spatial-temporal correlations between nodes in traffic flow forecasting. It is a critical task that obtaining the graph sequence models by connecting graph models from different time periods, in the field of graph theory \cite{47}. Particularly in the domain of traffic flow forecasting, the construction of effective dynamic graph models has emerged as a focal point of research. Multiview Spatial-Temporal Transformer Network \cite{is1} addresses the challenge of accurately predicting traffic flow by integrating multiple views for complex spatial-temporal feature learning. Although certain models demonstrate high accuracy in short-term forecasting, its performance significantly diminishes in long-term forecasting, thereby limiting its overall effectiveness. In such conditions, certain models may only be suitable for specific short-duration or long-duration tasks. Despite their overall utility, their performance in long-term forecasting tasks may still be notably poor, which is an important consideration in model selection and application. A key issue is the decline in accuracy for long-term forecasting. Spatial-Temporal Transformer Networks (STTN) \cite{b22} improve the accuracy of long-term flow forecasting by combining dynamic directional spatial dependencies and long-range temporal dependencies. Fang et al. \cite{b30} introduced a LSTM with attention mechanisms for short-term traffic flow forecasting, allocating weights to different inputs through the attention mechanism to focus on key information and improve forecasting accuracy. M{\'e}ndez et al. \cite{b31} presented a hybrid model combining CNNs and Bi-directional LSTMs to address long-term traffic flow forecasting issues. Djenouri et al. \cite{b32} explored an innovative graph convolutional neural network for urban traffic flow forecasting in edge IoT environments, employing a combined approach of graph optimization and forecasting.
	
	In summary, researchers have developed various advanced spatial-temporal models that utilize techniques such as GNNs, RNNs, attention mechanisms, and Transformers to capture the complex spatial and temporal dependencies and correlations within traffic networks. Some models also incorporate external factors such as weather and road conditions to enhance the accuracy of traffic flow forecasting. Previous work shows that obtaining both static and dynamic graph structures is crucial, and forecasting accuracy significantly declines with increased forecasting extent. Especially when considering urban traffic and long-term forecasting simultaneously, the accuracy of these models may decrease. Previous research mainly focused on graph structures and short-term or long-term forecasting, whereas this paper considers both static and dynamic spatial dependencies and introduces a multi-channel mechanism, combined with the characteristics of Transformers, aiming to ensure the accuracy of short-term forecasting while effectively capturing long-term temporal dependencies. 
	\section{Proposed Model}
	In this section, we describe MC-STTM in detail. First, we introduce some necessary symbols and definitions that will be used throughout the paper. Then, we describe the overall framework of our model, which is mainly composed of four parts: multi-channel mechanism, spatial-temporal block, feature fusion layer, and prediction layer. Afterwards, we elaborate on the main innovative parts of the framework: multi-channel mechanism, spatial block, and temporal Transformer.
	\subsection{Problem Definition}
	The topological structure of the road network is represented as a graph $\mathcal{G}=\left(\mathcal{V},\Theta,\mathbf{A}\right)$, where $\mathcal{V} = \{\mathcal{V}_1, \mathcal{V}_2, \cdots, \mathcal{V}_M\}$ denotes the set of nodes corresponding to $M$ roads or sensors in the road network. The set of edges is denoted by $\Theta$, and the spatial adjacency matrix $\mathbf{A}\in\mathbb{R}^{M\times M}$ contains the connectivity of the roads in the network (e.g., computed based on the Euclidean distance between two points). 
	
	$\mathbf{X}_{hour}$ and $\mathbf{X}_{day}$ represent two channels, which are selected from historical time slices at different time intervals. $\mathbf{X}_{hour}=\left[\mathbf{X}^{t-p+1}{,\mathbf{X}}^{t-p+2},\cdots,\mathbf{X}^t\right]$ represents recent traffic conditions, which exhibits temporal correlation over a short period of time;
	$\mathbf{X}_{day}=\left[\mathbf{X}^{t-ds}{,\mathbf{X}}^{t-(d-1)s},\cdots,\mathbf{X}^{t-s}\right]$ represents long-term traffic conditions, reflecting temporal correlation over a longer period of time; $\mathbf{Y}=\left[\mathbf{X}^{t+1},\mathbf{X}^{t+2},\cdots,\mathbf{X}^{t+q}\right]$ represents future traffic conditions; $p$ represents the number of historical time slices; $d$ represents the number of historical days; $s$ represents the number of time slices in a day (if the time interval of each time slice is 5 minutes, then $s=288$); $q$ represents the prediction horizon, where $q=1$ for single-step prediction and $q>1$ for multi-step prediction. 
	
	The task of traffic flow forecasting can be defined as follows: given the historical traffic conditions $\mathbf{X}_{hour}$ and $\mathbf{X}_{day}$, and a traffic road network topology $\mathcal{G}$, learn a prediction model $\mathcal{F}$ to forecast $\mathbf{Y}$. The expression can be represented as Eq. \eqref{equation1}:
	\begin{equation}
		\mathbf{Y}=\mathcal{F}\left(\mathcal{G};\mathbf{X}_{hour};\mathbf{X}_{day}\right).\label{equation1}
	\end{equation}
	\subsection{Model Framework}
	\begin{figure*}
		\centerline{\includegraphics[width=1\textwidth]{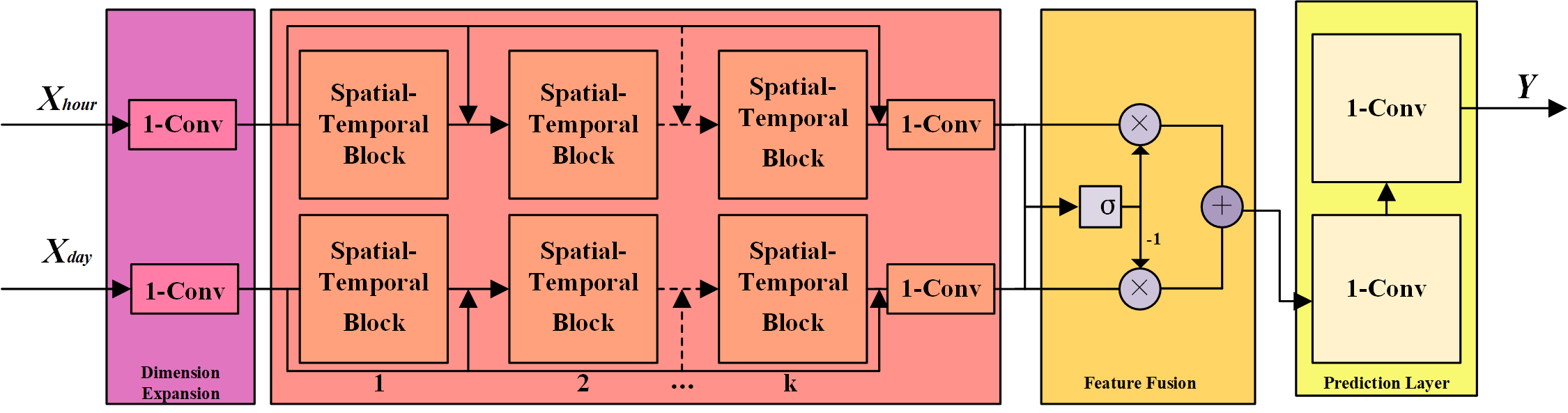}}
		\caption{The framework of MC-STTM.}
		\label{fig2}
	\end{figure*}
	As depicted in Fig. \ref{fig2}, MC-STTM adopts a multi-channel mechanism for input, where the features of the input are first dimensionally expanded and then subjected to stacked spatial-temporal blocks for feature extraction. Subsequently, the feature fusion block is employed, followed by the prediction layer, which yields the final prediction result. The dimensional expansion of the features can capture more information and relationships, thereby improving the model's performance and generalization ability. Each spatial-temporal block consists of a spatial block and a temporal block that jointly extract spatial-temporal features. The stacking of spatial-temporal blocks forms a deep model, which facilitates better feature extraction. Furthermore, the feature fusion adopts a gate mechanism similar to GRU that not only prevents gradient vanishing and explosion but also facilitates learning long-term dependencies. Finally, the prediction layer aggregates spatial-temporal features through two 1D convolution operations, enabling traffic flow forecasting.
	\paragraph{Multi-Channel Mechanism}
	The model learns from historical data and the topological structure of the road network $\mathcal{G}$ to predict future traffic conditions, which represents a single-channel prediction process. According to Eq. \eqref{equation1}, the so-called multi-channel refers to learning the model from historical data at multiple scales and the topological structure of the road network $\mathcal{G}$ to predict future traffic conditions. We can select historical data at different intervals, such as time slices ranging from a few minutes to several days or even weeks. This mechanism can learn more features and improve the performance of the model. We select $\mathbf{X}_{hour}$ and $\mathbf{X}_{day}$, where $\mathbf{X}_{hour}$ is characterized by short-term temporal correlations while $\mathbf{X}_{day}$ displays long-term temporal correlations.
	\begin{figure*}
		\centerline{\includegraphics[width=1\textwidth]{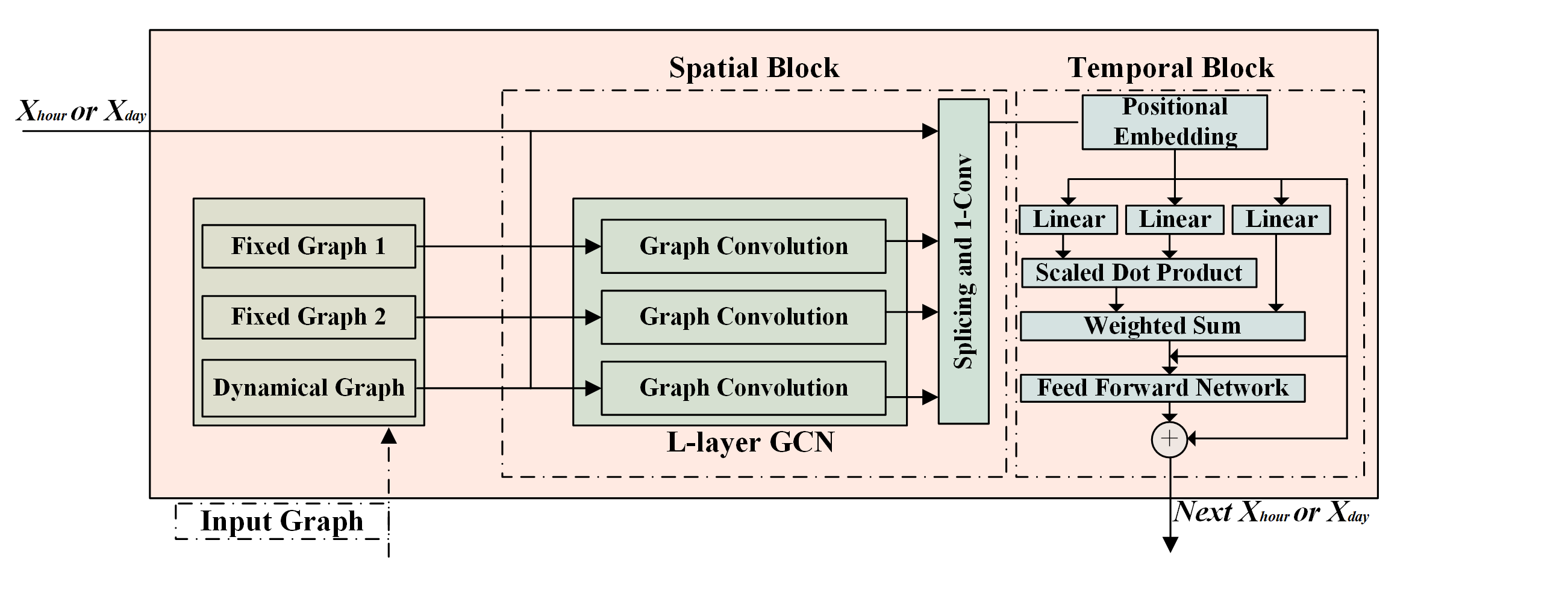}}
		\caption{The architrctaure of spatial-temporal block.}
		\label{fig3}
	\end{figure*}
	\paragraph{Spatial-Temporal Block}
	The future traffic condition of each node in a road network is influenced by multiple factors, such as the traffic conditions of adjacent or similar nodes, historical and forecast time steps (i.e., $p$ and $q$), and special events (e.g., traffic accidents and extreme weather conditions). We propose a spatial-temporal block that not only uses the spatial feature extractor GCN but also incorporates the Transformer, which show good performance in long-term prediction. As shown in Fig.~\ref{fig2}, multiple spatial-temporal blocks are stacked and connected by residual connections. The input to the \emph{k}-th spatial-temporal block is $\mathbf{X}^k$ ($\mathbf{X}_{hour}^k\in\mathbb{R}^{M\times p\times f_d}$ or $\mathbf{X}_{day}^k\in\mathbb{R}^{M\times d\times f_d})$ and $\mathcal{G}_k$, a 3D tensor with three dimensions representing the number of nodes $M$, the length of historical time slices $p$ or $d$, and the size of the feature dimension after expansion $f_d$. The output of the \emph{k}-th spatial-temporal block is $\mathbf{X}^{k+1}$ ($\mathbf{X}_{hour}^{k+1}\in\mathbb{R}^{M\times p\times f_d}$ or $\mathbf{X}_{day}^{k+1}\in\mathbb{R}^{M\times d\times f_d}$). In fact, the input of the next spatial-temporal block is jointly composed of the input and output of the previous spatial-temporal block. To simplify the expression, this is not explicitly reflected in the formula but can be clearly observed in Fig.~\ref{fig2}.
	\begin{equation}
		\mathbf{X}^{k+1}=\mathcal{F}_k\left(\mathcal{G}_k;\mathbf{X}^k\right).\label{equation2}
	\end{equation}
	
	Therefore, we can infer that the output of the final spatial-temporal block is given by $\mathbf{X}^{ST}=Conv{(\mathbf{X}}^k+\mathbf{X}^{k+1})$, where $\mathbf{X}^{ST}$ represent either $\mathbf{X}_{hour}^{ST}\in\mathbb{R}^{M\times p\times f_ d}$ or $\mathbf{X}_{day}^{ST}\in\mathbb{R}^{M\times d\times f_ d}$. Finally, to achieve dimension alignment between $\mathbf{X}_{hour}^{ST}$ and $\mathbf{X}_{day}^{ST}$, they are separately processed through a 1D convolution, resulting in $\hat{\mathbf{X}}_{hour}^{ST}\in\mathbb{R}^{M\times p\times f_ d}$ and $\hat{\mathbf{X}}_{day}^{ST}\in\mathbb{R}^{M\times p\times f_ d}$.
	
	\paragraph{Feature Fusion}
	A gate mechanism similar to GRU is utilized to fuse $\hat{\mathbf{X}}_{hour}^{ST}$ and $\hat{\mathbf{X}}_{day}^{ST}$, with the gate $g$ calculated from both inputs.
	\begin{equation}
		g=sigmoid\left(f_1\left(\hat{\mathbf{X}}_{hour}^{ST}\right)+f_2\left(\hat{\mathbf{X}}_{day}^{ST}\right)\right),\label{equation3}
	\end{equation}
	where $f_1$ and $f_2$ denote linear transformations that convert $\hat{\mathbf{X}}_{hour}^{ST}$ and $\hat{\mathbf{X}}_{day}^{ST}$ into 1D vectors, which are then passed through a sigmoid function. 
	\begin{equation}
		{\hat{\mathbf{X}}}_{gate}^{ST}=g\hat{\mathbf{X}}_{hour}^{ST}+\left(1-g\right)\hat{\mathbf{X}}_{day}^{ST}.\label{equation4}
	\end{equation}
	
	The output ${\hat{\mathbf{X}}}_{gate}^{ST}\in\mathbb{R}^{M\times p\times f_d}$ from the feature aggregation layer is used as input for the final prediction layer to generate the ultimate result.
	\paragraph{Prediction Layer}
	The prediction layer consists of two 1D convolutions, which perform single-step or multi-step forecasting on ${\hat{\mathbf{X}}}_{gate}^{ST}$ based on $p$ and $q$, and the features of ${\hat{\mathbf{X}}}_{gate}^{ST}$ are dimensionally reduced from $f_d$ to 1. The final predicted result is $\mathbf{Y}\in\mathbb{R}^{M\times q\times1}$.
	\begin{equation}
		\mathbf{Y}=Conv\left(Conv\left({\hat{\mathbf{X}}}_{gate}^{ST}\right)\right).\label{equation5}
	\end{equation}
	\subsection{Multi-Channel Mechanism}
	In this study, a multi-channel mechanism is employed to decompose historical traffic data into recent patterns and daily patterns. The recent pattern is obtained by collecting the traffic flow at a frequency of $p$ times per day, and the resulting time series is denoted as $\left[\mathbf{X}^{t-p+1}{,\mathbf{X}}^{t-p+2},\cdots,\mathbf{X}^t\right]$. The daily pattern is obtained by considering the data from the previous $d$ days, and the resulting time series is denoted as $\left[\mathbf{X}^{t-ds}{,\mathbf{X}}^{t-(d-1)s},\cdots,\mathbf{X}^{t-s}\right]$, the parameter $s$ represents the number of time slices within a day, and if the time interval is 5 minutes, the value of $s$ is 288. Simultaneously, if $p=12$, $d=7$, then the recent pattern reflected the continuous traffic conditions within the past hour, while the daily pattern considered the daily cyclic components in the traffic data (e.g., peak hours in the morning and evening, fixed school commute times near schools, and fixed peak travel schedules near transportation hubs).
	
	\subsection{Spatial Block}
	The GCN is utilized in the current task to extract complex topological information from road networks, Given the structural information of road network nodes, GCN serves as a fundamental approach for extracting node features. A two-layered GCN can be formulated as Eq. \eqref{equation6} as follows:
	\begin{equation}
		\left\{\begin{matrix}\mathbf{H}_1=\sigma\left(\mathbf{AX}\mathbf{W}_0^H\right)\\\mathbf{H}_2=\sigma(\mathbf{A}\mathbf{H}_1\mathbf{W}_1^H)\\\end{matrix}\right.,\label{equation6}
	\end{equation}
	where $\sigma$ represents the activation function, typically ReLU or tanh. $\mathbf{X}\in\mathbb{R}^{M\times D}$ denotes the node feature matrix. $\mathbf{W}_0^H\in\mathbb{R}^{D\times D}$ and $\mathbf{W}_1^H\in\mathbb{R}^{D\times D}$ denote the model parameter. Matrix $\mathbf{A}\in\mathbb{R}^{M\times M}$ refers to the normalized adjacency matrix, and $\mathbf{H}_1\in\mathbb{R}^{N\times D}$ denotes the output of the first layer. The ultimate output is a node feature representation matrix $\mathbf{H}_2$, based on the given structural information of the road network nodes.
	
	An adaptive adjacency matrix $\mathbf{A}_{adp}$, similar to that in \cite{b25}, is introduced. This adaptive adjacency matrix does not require any prior knowledge and is learned through gradient descent. By using this model, the model can learn the hidden relationships between nodes within the road network. Column vectors $\mathbf{E}_c$ and row vectors $\mathbf{E}_r$ with learnable parameters are randomly initialized to generate the adaptive adjacency matrix as follows:
	\begin{equation}
		\mathbf{A}_{adp}=softmax\left(ReLU\left(\mathbf{E}_c\mathbf{E}_r\right)\right),\label{equation7}
	\end{equation}
	where ReLU is used to eliminate weakly correlated nodes and softmax function is applied for normalization.
	
	We propose the following graph convolution layer:
	\begin{equation}
		{\hat{\mathbf{X}}}^k=\mathbf{A}_{adp}\mathbf{X}^k\mathbf{W}_0^k+\mathbf{A}_{fixed}^1\mathbf{X}^k\mathbf{W}_1^k+\mathbf{A}_{fixed}^2\mathbf{X}^k\mathbf{W}_2^k,\label{equation8}
	\end{equation}
	where ${\hat{\mathbf{X}}}^k$ represent either ${\hat{\mathbf{X}}}^k_{hour}$ or ${\hat{\mathbf{X}}}^k_{day}$.The adjacency matrices $\mathbf{A}_{fixed}^1$ and $\mathbf{A}_{fixed}^2$ are built based on the physical distances between nodes in opposite directions. The correlation between nodes decreases as their physical distance increases. Eq. \eqref{equation8} allows us to extract not only the static spatial features of road networks from fixed graphs, but also the dynamic spatial features using an adaptive adjacency matrix. This approach enhances the accuracy of feature extraction for road networks.
	
	\subsection{Temporal Transformer}
	As shown in Fig.~\ref{fig3}, a Transformer-based temporal block is presented for capturing long-term temporal dependencies, which outperforms traditional RNNs and their variants. Firstly, position encoding is applied to each time step. Differently from previous work, we perform position encoding not only for all time slices within a day but also for weekdays and weekends within a week. After position encoding, the feature vectors in the \emph{k}-th ST block are mapped to query vector $\mathbf{Q}^k$, key vector $\mathbf{K}^k$, and value vector $\mathbf{V}^k$ by three learned linear transformations, similar to the original Transformer. 
	\begin{equation}
		\begin{matrix}\mathbf{Q}^k={\acute{\mathbf{X}}}^k\mathbf{W}_Q^k\\\mathbf{K}^k={\acute{\mathbf{X}}}^k\mathbf{W}_K^k\\\mathbf{V}^k={\acute{\mathbf{X}}}^k\mathbf{W}_V^k\\\end{matrix},\label{equation9}
	\end{equation}
	where ${\acute{\mathbf{X}}}^k$ represent either ${\acute{\mathbf{X}}}^k_{hour}$ or ${\acute{\mathbf{X}}}^k_{day}$. $\acute{\mathbf{X}}^k$ is obtained after position encoding $\hat{\mathbf{X}}^k$. The position encoding is performed separately for the 288 time slices within a day and the seven days of the week using one-hot encoding. $\mathbf{W}_Q^k$, $\mathbf{W}_K^k$, and $\mathbf{W}_V^k$ are learnable parameters. The softmax function is applied to convert the results into attention scores.
	\begin{equation}
		\mathbf{S}^k=softmax\left(\frac{\mathbf{Q}^k{\mathbf{K}^k}^T}{\sqrt{f_d}}\right).\label{equation10}
	\end{equation}
	
	After computing $\mathbf{S}^k$, we aggregate the result with $\mathbf{V}^k$ through dot-product attention and then pass it through a feed-forward neural network to produce the final output of the temporal block.
	\begin{equation}
		\mathbf{X}^{k+1}=ReLU\left(ReLU\left(\mathbf{M}^k\mathbf{W}_0\right)\mathbf{W}_1\right)\mathbf{W}_2+\mathbf{M}^k,\label{equation11}
	\end{equation}
	where $\mathbf{M}^k=\mathbf{S}^k\mathbf{V}^k+{\acute{\mathbf{X}}}^k$. This temporal feature extraction process differs from recurrent neural networks and their variants in that it can process sequence data in parallel.
	\section{Experiments}
	In this section, we conduct experiments on six real-world datasets to address the following questions. \textbf{Q1}: Does the performance of deep learning models improve before and after introducing multi-channel mechanisms? \textbf{Q2}: How does the performance of our model compare to other advanced traffic flow forecasting models? We also perform ablation experiments to demonstrate the importance of each block.
	\subsection{Experiment Setting}
	
	\begin{table*}[h]
		\caption{Dataset Description.}\label{tab1}%
		\resizebox{\textwidth}{!}{%
			\begin{tabular}{llllll}
				\hline
				\textbf{Dataset}&\textbf{Nodes}&\textbf{Time Slices}&\textbf{Edges}&\textbf{Time Range}&\textbf{The selected features} \\
				\hline
				PEMS03 \cite{b26}&358&26208&547&9/1/2018-11/30/2018&flow\\
				
				PEMS04 \cite{b26}&307&16992&340&1/1/2018-2/28/2018&flow\\
				
				PEMS07 \cite{b26}&883&28224&866&5/1/2017-8/31/2017&flow\\
				
				PEMS08 \cite{b26}&170&17856&295&7/1/2017-8/31/2017&flow\\
				
				METR-LA \cite{b27}&207&34272&1515&3/1/2012-6/30/2012&speed\\
				
				PEMS-BAY \cite{b27}&325&52116&2369&1/1/2017-5/31/2017&speed\\
				\hline
		\end{tabular}}
	\end{table*}
	\begin{figure*}
		\centerline{\includegraphics[width=0.5\textwidth]{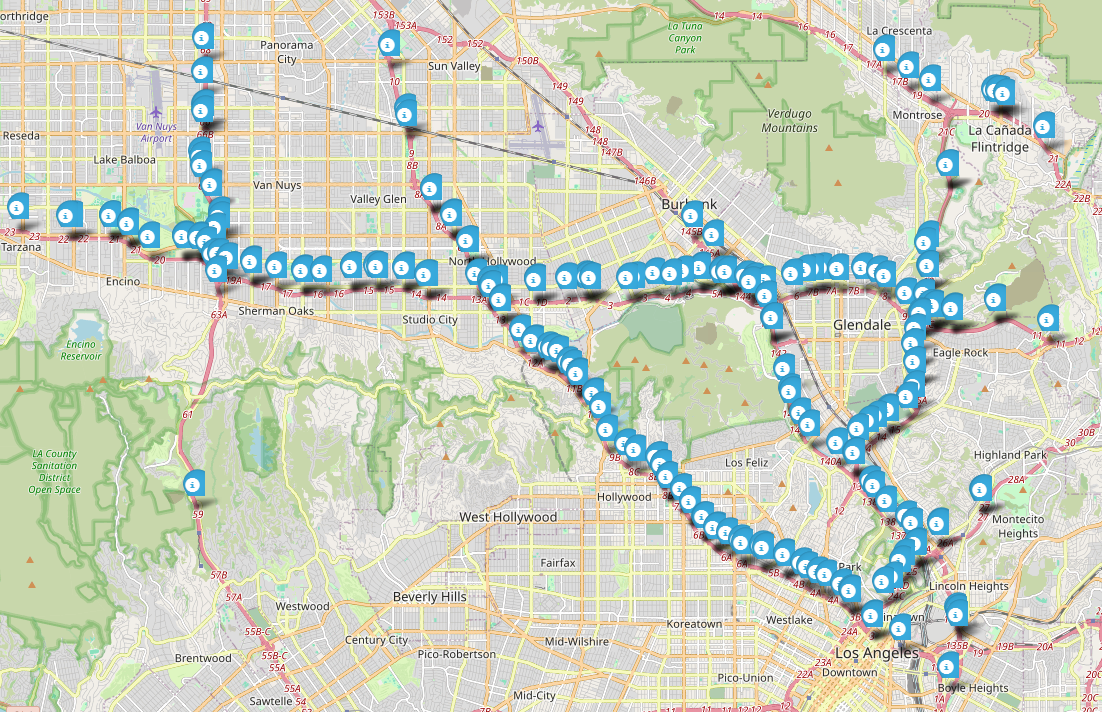}}
		\caption{Visualization of sensor distribution in the METR-LA dataset.}
		\label{fig333}
	\end{figure*}
	\paragraph{Dataset}
	In our study, we conduct a comprehensive performance evaluation of our method using real-world datasets from six different regions, as shown in Table \ref{tab1}. The datasets are counted at a 5-min interval. These datasets vary in aspects such as the number of nodes, time slices, edges (i.e., spatial distance relationships between nodes), and the duration across time. Specifically, we utilize traffic flow data for experiments in the PEMS03, PEMS04, PEMS07, and PEMS08 datasets, while traffic speed data is used in the METR-LA and PEMS-BAY datasets. This diverse dataset selection, along with Fig. \ref{fig333} illustrating the sensor distribution for METR-LA, provides an in-depth understanding of dataset's geographical and technical features, ensuring a thorough assessment of our model's applicability and effectiveness.
	\paragraph{Evalution metrics}
	The evaluation metrics used in this study are the Mean Absolute Error (MAE), Mean Absolute Percentage Error (MAPE), and Root Mean Square Error (RMSE).
	\paragraph{Implementation details}
	For the sake of fairness, we divide the datasets into training, validation, and test sets using the same ratio as the baseline, namely 6:2:2. Our objective is to predict the traffic flow for the next hour based on one hour of historical data. The hyperparameter settings for the baseline models are selected based on the default configurations provided in the publicly available code. This approach ensures consistency and reproducibility in our experimental evaluation. In the multi-channel mechanism, we employ one week's historical data from an additional channel. All experiments are conducted on the windows platform (CPU: Intel(R) Core(TM) i7-9700K CPU @ 3.60GHz; GPU: NVIDIA GeForce RTX 2080). We set the following hyperparameters: 4 heads for the multi-head attention mechanism, PyTorch as the deep learning framework, and MAE as the loss function. The maximum number of epochs is set to 200, with a learning rate of 0.0001 and a batch size of 64. Other parameters are kept at their default values. During the training process, we automatically save the best model on the validation set. If the model did not improve for 15 consecutive epochs, we consider it to converge and stop the training process.
	\paragraph{Baselines}
	We compare MC-STTM against the following popular approaches for traffic flow forecasting task:
	\begin{itemize}
		\item \textbf{HA}: this approach leverages a weighted mean of historical data slices to prognosticate forthcoming data points.
		\item \textbf{Temporal models}: RNN; LSTM; GRU; Transformer.
		\item \textbf{T-GCN} \cite{b6}: a spatial-temporal hybrid model that combines GCN and GRU. 
		\item \textbf{STGCN} \cite{b16}: it features a complete convolutional structure, enabling faster training speed with lower model complexity.
		\item \textbf{Graph WaveNet} \cite{b25}: it introduces an adaptive adjacency matrix, leading to significantly improved performance.
		\item \textbf{STTN} \cite{b22}: the model performs well in long-term prediction, but has relatively high model complexity.
	\end{itemize}
	\subsection{Performance Comparison before and after Introduction of Multi-Channel Mechanism}
\begin{figure}
		\centerline{\includegraphics[width=1\textwidth]{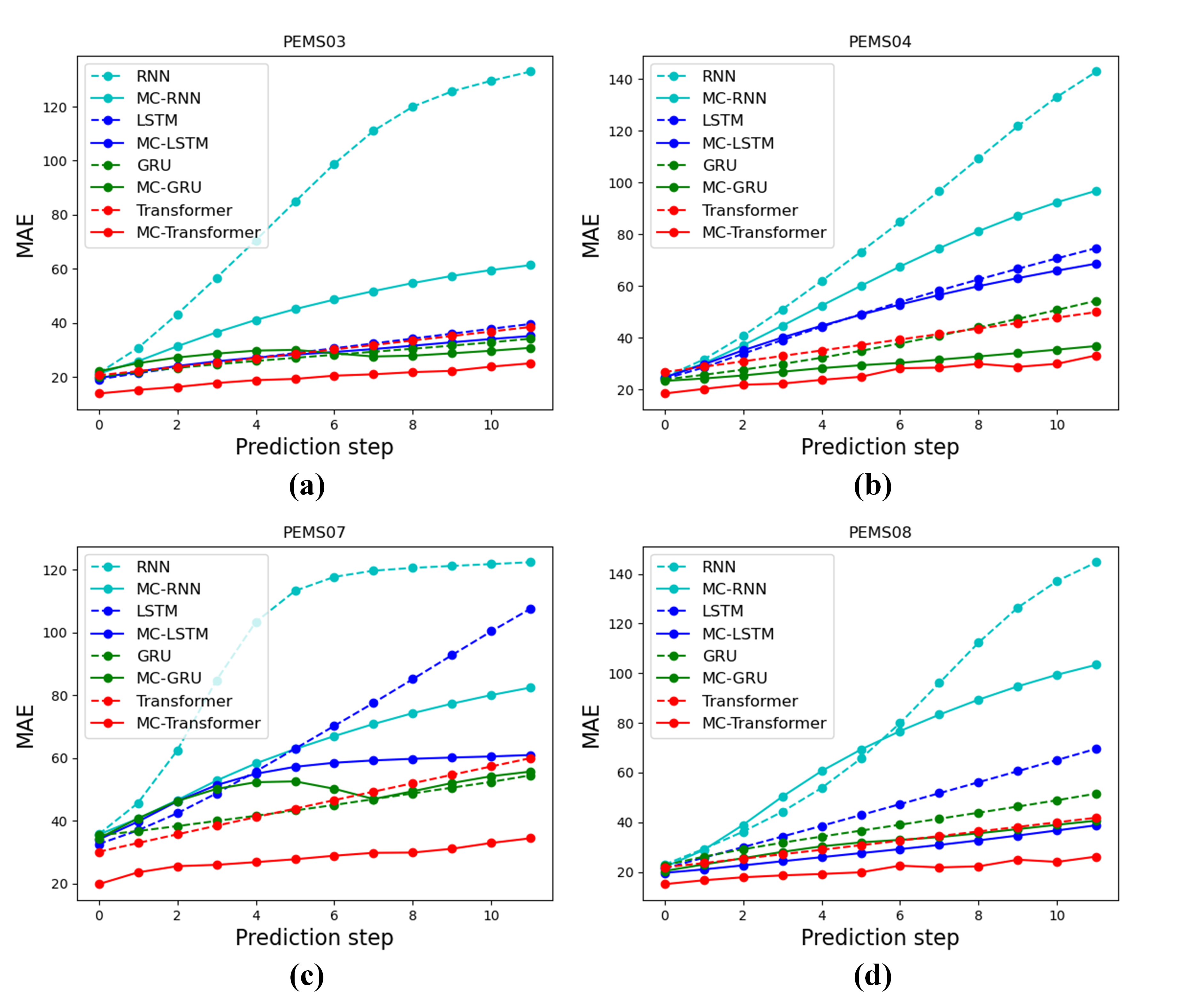}}
		\caption{MAE comparison of flow prediction before and after introducing multi-channel mechanism in temporal models: (a) results on PEMS03; (b) results on PEMS04; (c) results on PEMS07; (d) results on PEMS08.}
		\label{fig4}
	\end{figure}

\begin{figure}
		\centerline{\includegraphics[width=1\textwidth]{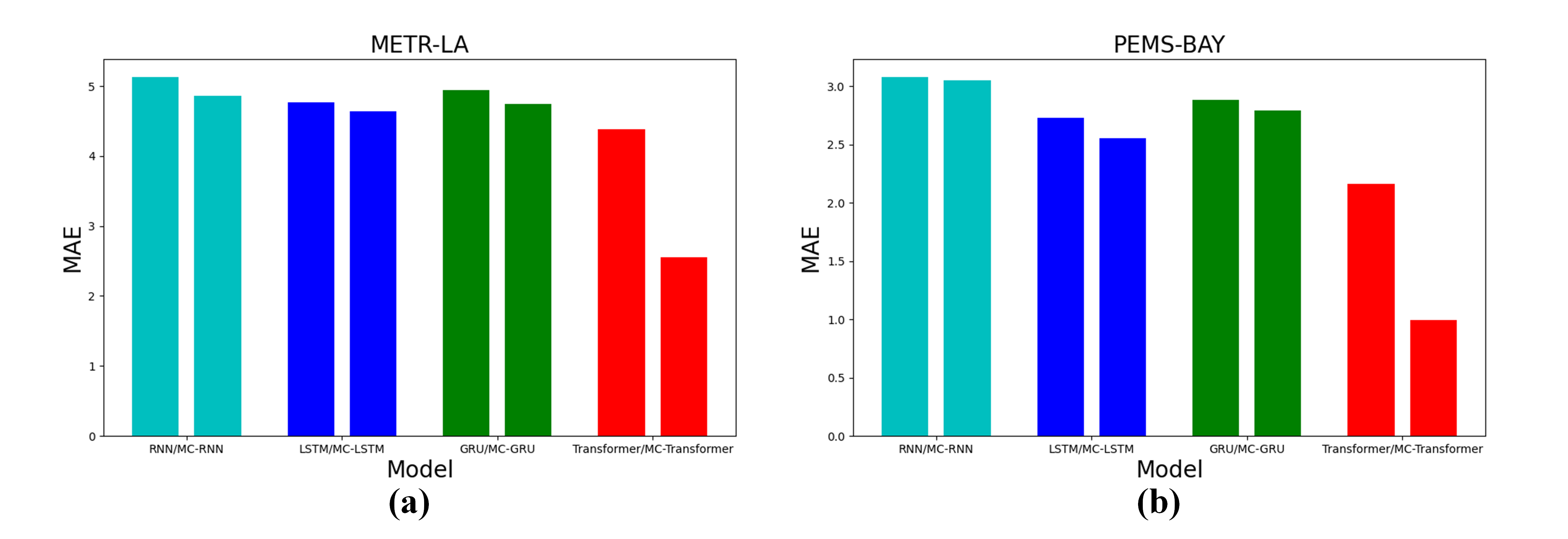}}
		\caption{MAE comparison of speed prediction before and after introducing multi-channel mechanism in temporal models: (a) results on METR-LA; (b) results on PEMS-BAY.}
		\label{fig66}
	\end{figure}

	To demonstrate that the performance of deep learning models can be improved by introducing multi-channel mechanisms, temporal models in baselines are employed to introduce multi-channel mechanisms on different datasets. The models prefixed with \textbf{MC} denote that they incorporate multi-channel mechanisms. Fig. \ref{fig4} depicts the comparison of MAE in four real-world flow datasets, before and after incorporating the multi-channel mechanism in the each temporal model, the solid line represents the performance curve with the multi-channel mechanism, which exhibits noticeably lower MAE values than the dashed line. Furthermore, this improvement in flow datasets is particularly prominent in long-term forecasting. We also conduct comparative experiments on two real speed datasets. As shown in Fig. \ref{fig66}, we compare the results of single-step prediction before and after introducing the multi-channel mechanism in the METR-LA and PEMS-BAY datasets. The results indicate that the multi-channel mechanism improved the model performance in the speed datasets. Moreover, the improvement in performance is more significant in Transformer-based models compared to RNN-based models, which is one of the reasons for selecting the Transformer network architecture in this paper.
	
	The results indicate that the introduction of a multi-channel mechanism improves the prediction performance of deep learning models across different datasets, thus answering our research question \textbf{Q1}: the performance of deep learning models will be significantly improved by introducing a multi-channel mechanism.
	\subsection{Performance Comparison between MC-STTM and Baseline Model}
	To answer \textbf{Q2}, we compare the traffic flow forecasting performance of MC-STTM with baseline models on six different datasets. Currently, most mainstream traffic flow forecasting models are spatial-temporal approaches. Building upon existing research, MC-STTM innovates not only in terms of multi-channel representation but also integrates both static and dynamic spatial dependencies as well as long-term temporal dependencies. The experiment is a multi-step prediction, with steps 1 to 12 ranging from 5 minutes to 60 minutes. Table \ref{tab6} and \ref{tab99} show a performance comparison between MC-STTM and baseline models, where bold indicates the best results and underlined indicates the second-best.
	\begin{table*}[h]
		\caption{Performance comparison of flow prediction between MC-STTM and other baseline models.}\label{tab6}%
		\resizebox{\textwidth}{!}{%
			\begin{tabular}{lllllllllll}
				\hline
				\multirow{2}{*}{\textbf{Dataset}} & \multirow{2}{*}{\textbf{Model}} & \multicolumn{3}{c}{\textbf{\textit{15min (step 3)}}}& \multicolumn{3}{c}{\textbf{\textit{30min (step 6)}}}& \multicolumn{3}{c}{\textbf{\textit{60min (step 12)}}} \\
				\cmidrule{3-11}
				& & MAE & MAPE & RMSE & MAE & MAPE & RMSE & MAE & MAPE & RMSE \\
				\hline
				\multirow{7}{*}{\textbf{PEMS03}} & HA & 31.89 & 25.82 & 43.28 & 31.89 & 25.82 & 43.28& 31.89 & 25.82 & 43.28\\
				
				& GRU & 22.02 & 20.26 & 35.54 & 24.73 & 22.25 & 38.74 & 31.60 & 26.21 & 47.20\\
				
				& TGCN & 17.24 & 16.33 & \underline{25.36} & 20.42 & 20.30 & \underline{29.80} & 26.76 & 29.25 & 38.12\\
				
				& STGCN & 17.69 & 14.85 & 30.29 & 20.94 & 17.41 & 34.25 & 26.78 & \underline{21.90} & 42.48\\
				
				& Graph WaveNet &\underline{17.17} &\underline{14.55} &25.76 &20.46 &\underline{16.98} &30.95 &29.18 &23.94 &43.36\\
				
				& STTN & 19.18&18.84 &28.70 &\underline{20.29} &19.38 &30.41 & \underline{24.39}& 23.01&\underline{35.94}\\
				
				& MC-STTM & \textbf{16.56} & \textbf{14.07} & \textbf{24.74} & \textbf{18.35} & \textbf{15.21} & \textbf{27.53} & \textbf{22.30} &\textbf{17.88} & \textbf{33.26}\\
				\hline
				\multirow{7}{*}{\textbf{PEMS04}}& HA & 42.28 & 30.44 & 57.69 & 42.28 & 30.44 & 57.69 & 42.28 & 30.44 & 57.69\\
				
				& GRU & 24.35 & 23.91 & 37.76 & 26.82 & 26.60 & 40.51 & 34.22 & 38.22 & 49.38\\
				
				& TGCN & 21.20 & 15.76 & 31.65 & 25.67 & 20.49 & 37.37 & 34.91 & 31.22 & 49.02\\
				
				& STGCN & 21.69 & \underline{15.41} & 32.43 & 26.45 & 18.64 & 39.39 & 35.69 & 24.33 & 54.44\\
				
				& Graph WaveNet &22.44 &15.78 &33.79 &26.64 &18.98 &39.78 &37.95 &27.48 &54.74\\
				
				& STTN & \underline{20.87} & \textbf{15.36} & \underline{31.28} & \underline{22.83} & \textbf{16.88} & \underline{34.36} & \underline{26.80} & \textbf{20.39} & \underline{39.98}\\
				
				& MC-STTM & \textbf{19.74} & 15.47 & \textbf{29.65} & \textbf{21.79} & \underline{17.26} & \textbf{32.57} & \textbf{25.63} & \underline{20.90} & \textbf{37.59}\\
				\hline
				\multirow{7}{*}{\textbf{PEMS07}} & HA & 48.95 & 21.40 & 64.91 & 48.95 & 21.40 & 64.91 & 48.95 & 21.40 & 64.91\\
				
				& GRU & 35.45 & 19.93 & 58.82 & 39.84 & 22.33 & 62.87 & 49.04 & 27.16 & 72.57\\
				
				& TGCN & 25.50 & 11.51 & 37.70 & 34.34 & 16.84 & 49.54 & 49.81 & 27.15 & 67.75\\
				
				& STGCN & 25.70 & 10.81 & 39.07 & 32.16 & 14.15 & 49.48 & 43.13 & 20.44 & 67.41\\
				
				& Graph WaveNet &25.07 &10.42 &37.99 &29.97 &12.81 &44.86 &43.51 &20.29 &60.73\\
				
				& STTN & \underline{23.52}& \underline{10.27}&\underline{34.83} &\underline{26.26} &\underline{11.50} &\underline{38.65} &\underline{32.24} &\underline{14.65} &\underline{46.14}\\
				
				& MC-STTM & \textbf{20.75} & \textbf{8.65} & \textbf{31.75} & \textbf{23.29} & \textbf{9.82} & \textbf{35.47} & \textbf{28.18} & \textbf{12.31} & \textbf{42.07}\\
				\hline
				\multirow{7}{*}{\textbf{PEMS08}} & HA &34.09 &19.59 &45.58 &34.09 &19.59 &45.58 &34.09 &19.59 &45.58\\
				
				& GRU & 23.83 & 13.78 & 35.08 & 27.30 & 15.50 & 39.51 & 34.87 & 20.69 & 20.04\\
				
				& TGCN & 17.78 & 10.99 & 25.81 & 22.31 & 14.54 & 31.76 & 31.85 & 22.87 & 43.64\\
				
				& STGCN & \underline{16.96} & \underline{9.65} & \underline{25.27} & 21.23 & 12.06 & 31.92 & 29.04 & 16.17 & 42.06\\
				
				& Graph WaveNet &17.79 &10.34 &26.74 &21.45 &12.30 &32.03 &31.04 &18.15 &43.75\\
				
				& STTN &18.28 &10.41 &26.66 &\underline{19.04} &\underline{10.87} &\underline{28.20} &\underline{22.60} &\underline{12.92} &\underline{33.47}\\
				
				& MC-STTM & \textbf{15.86} & \textbf{9.03} & \textbf{23.48} & \textbf{17.19} & \textbf{9.92} & \textbf{25.73} & \textbf{20.11} & \textbf{10.04} & \textbf{25.98}\\
				\hline
		\end{tabular}}
	\end{table*}
	
\begin{figure*}
		\centerline{\includegraphics[width=1\textwidth]{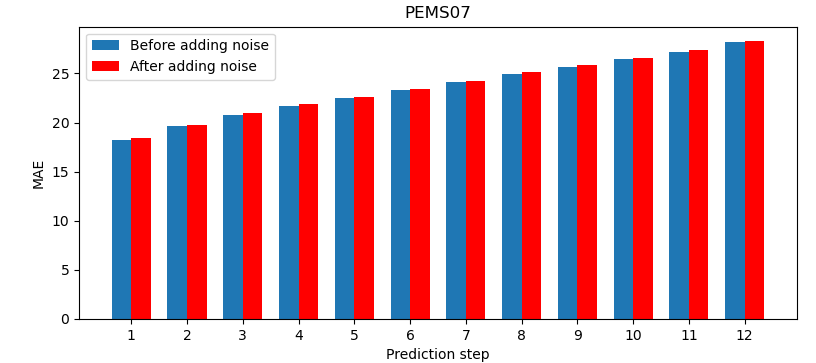}}
		\caption{Visualization of performance comparison between before and after adding Gaussian noise in the PEMS07 dataset.}
		\label{figis}
	\end{figure*}

	\begin{table*}[h]
		\caption{Performance comparison of speed prediction between MC-STTM and other baseline models.}\label{tab99}%
		\resizebox{\textwidth}{!}{%
			\begin{tabular}{lllllllllll}
				\hline
				\multirow{2}{*}{\textbf{Dataset}} & \multirow{2}{*}{\textbf{Model}} & \multicolumn{2}{c}{\textbf{\textit{15min (step 3)}}}& \multicolumn{2}{c}{\textbf{\textit{30min (step 6)}}}& \multicolumn{2}{c}{\textbf{\textit{60min (step 12)}}} \\
				\cmidrule{3-8}
				& & MAE &  RMSE & MAE &  RMSE & MAE &  RMSE \\
				\hline
				\multirow{7}{*}{\textbf{METR-LA}} & HA & 7.41 & 12.77 & 7.41 &  12.77 & 7.41 &  12.77 \\
				
				& GRU &5.82&11.76&8.12&14.63&11.24&18.43\\
				
				& TGCN & 4.09 &  7.91 & 5.37 & 9.93 & 7.41 &12.57\\
				
				& STGCN&6.02&11.50&8.67&15.46&9.57&17.03\\
				
				& Graph WaveNet&\textbf{3.64}&\underline{7.86}&\underline{4.78}&10.00&6.74&12.62\\
				
				& STTN &\underline{3.69}&8.06&4.80&\underline{9.80} &\underline{6.43} &\underline{12.22}\\
				
				& MC-STTM&3.79&\textbf{7.53}&\textbf{4.77}&\textbf{9.35}&\textbf{6.39}&\textbf{11.72}\\
				\hline
				\multirow{7}{*}{\textbf{PEMS-BAY}}& HA & 3.10 &   5.79 & 3.10 &   5.79 & 3.10 &   5.79\\
				
				& GRU&5.79&11.79&8.37&15.60&10.91&19.53\\
				
				& TGCN&1.82&3.23&2.45&4.31& 3.36&5.75\\
				
				& STGCN&2.44&4.30&2.81&5.05&3.51&6.50\\
				
				& Graph WaveNet&1.54&3.06&2.13&4.30&2.98&5.74\\
				
				& STTN&\underline{1.51}&\underline{2.84}& \underline{1.99}&\textbf{3.63}&\underline{2.51}&\textbf{4.57}\\
				
				& MC-STTM&\textbf{1.48}&\textbf{2.75}&\textbf{1.94}&\underline{3.64}&\textbf{2.45}&\underline{4.58}\\
				\hline
		\end{tabular}}
	\end{table*}
	The results indicate that MC-STTM outperforms baseline models on various datasets, particularly exhibiting superior performance in long-term prediction. This validates the superiority of the Transformer in capturing temporal features over extended periods. 
	Traffic flow prediction needs to account for external factors such as weather, road conditions, or special events that can impact traffic flow. However, due to limited information in the dataset, we conduct robustness experiments specifically on the PEMS07 dataset, which has the highest number of road network nodes. In the test set, we introduce Gaussian noise with a mean of 0 and a standard deviation of 0.01 as interference. The experimental results before and after adding noise are shown in Fig. \ref{figis}. The results indicate that there is little difference in traffic flow predictions before and after adding noise. Therefore, MC-STTM demonstrates strong interference resistance capabilities.
	\subsection{Ablation Experiments}
	In order to analyze the necessity of each block in MC-STTM, ablation experiments are conducted in this section. As shown in Table \ref{tab7}, each component contributes to the improvement of the model's performance. The performance metrics in the table represent the average values within one hour, aiming to evaluate the overall effectiveness of the model in multi-step prediction. Therefore, when modeling traffic data, it is necessary to consider both the spatial scale and the temporal scale, namely the S-Block and T-Block. At the spatial scale, the existing topological structure is undoubtedly important, but it is also essential to adopt dynamic modeling methods such as adaptive adjacency matrices to capture potential spatial correlations. Finally, the multi-channel mechanism is one of the main innovations in this paper, playing a vital role in capturing long-term dependencies.
	\begin{table*}[h]
		\caption{Results of the Ablation Experiments.}\label{tab7}%
		\resizebox{\textwidth}{!}{%
			\begin{tabular}{lllllllllllll}
				\hline
				\multirow{2}{*}{} & \multicolumn{3}{c}{\textbf{\textit{PEMS03}}}& \multicolumn{3}{c}{\textbf{\textit{PEMS04}}}& \multicolumn{3}{c}{\textbf{\textit{PEMS07}}}& \multicolumn{3}{c}{\textbf{\textit{PEMS08}}} \\
				
				& MAE & MAPE & RMSE & MAE & MAPE & RMSE & MAE & MAPE & RMSE& MAE & MAPE & RMSE \\
				\hline
				No $A_{adp}$ &21.89 &18.21 &32.94 &24.52 & 19.70&36.33 &26.44 &11.59 &39.69 &19.34 &11.35 &28.85\\
				
				No Fixed Graph &21.71 &19.34 &32.55 &29.60 &30.20 & 41.61&24.50 &10.58 &36.89 &19.63 & 11.53&28.83\\
				
				No S-Block &19.72 &16.78 &29.11 &23.82 &17.75 &35.44 &25.07 &10.67 &37.97 & 18.34&10.49 &27.25\\
				
				No T-Block&19.50 &16.48&29.37 &26.40 &20.43 &38.84 &26.94 &12.23 &39.25 &19.88 &12.02 &29.17\\
				
				No Multi Channel& 19.40 & 18.29 & 28.81 & 24.84 & 22.26 & 36.01 & 26.12 & 11.39 & 38.96 & 21.16 & 13.23 & 30.37\\
				
				\textbf{Full Model}& \textbf{18.63} & \textbf{15.38} & \textbf{28.11} & \textbf{21.97} & \textbf{17.59} & \textbf{32.88} & \textbf{23.55} & \textbf{10.02} & \textbf{35.99} & \textbf{17.38} & \textbf{10.03} & \textbf{25.98}\\
				\hline
		\end{tabular}}
	\end{table*}
	\begin{figure*}
		\centerline{\includegraphics[width=1\textwidth]{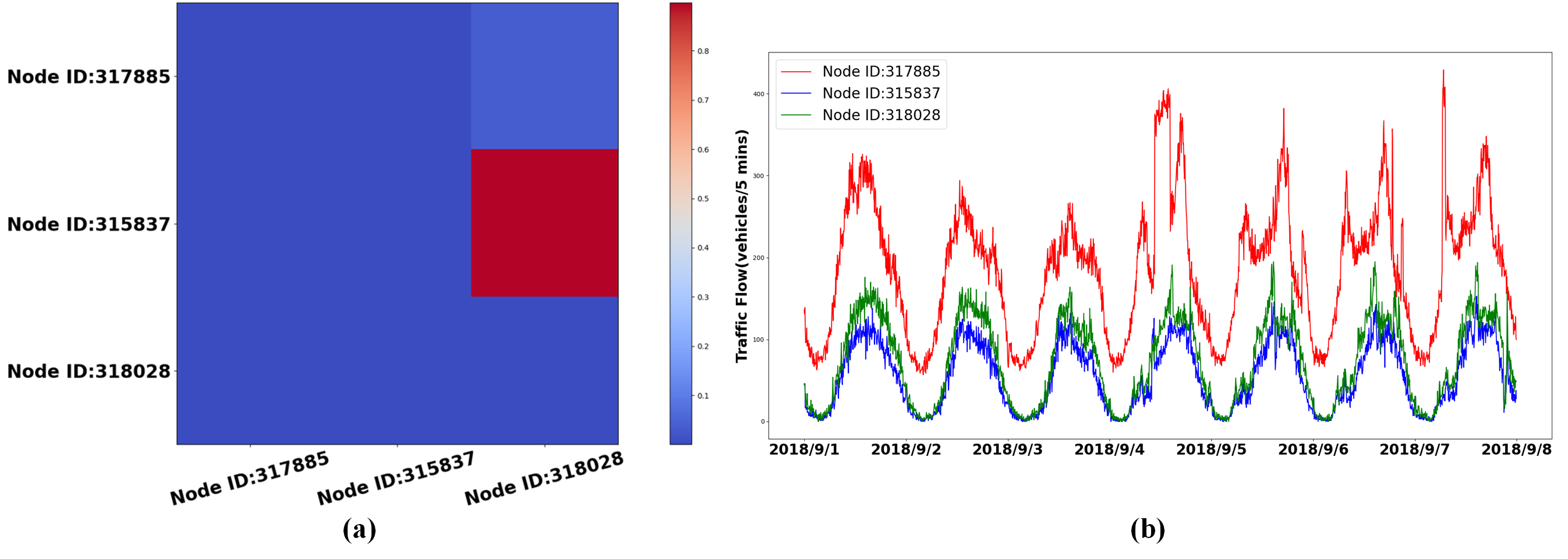}}
		\caption{Visualization of correlation between nodes: \textbf{(a)} partial results of the adaptive adjacency matrix in the PEMS03 dataset; \textbf{(b)} flow curves of corresponding nodes in the PEMS03 dataset.}
		\label{fig5}
	\end{figure*}
	
	The adaptive adjacency matrix, as generated by Equation 7, comprises values that are learnable parameters of the model. These values are dynamically optimized with each training iteration to discover the optimal configuration. This adaptive approach ensures that the adjacency matrix continually adapts and refines its parameters throughout the training process, enhancing the model's ability to capture complex relationships and dependencies in the data. In PEMS03, three known nodes are selected that are not directly connected in the road network. Through the learning of an adaptive adjacency matrix, we can observe from Fig. \ref{fig5}(a) that node ID 318028 exhibits a strong correlation with node ID 315837, while showing a relatively weak correlation with node ID 317885. Correspondingly, Fig. \ref{fig5}(b) reveals the traffic flow variations of these three nodes over the course of a week, which clearly align with the learned correlations from the adjacency matrix. The adaptive adjacency matrix not only captures the static spatial relationships present in the original road network, but also learns dynamic correlations. Static spatial relationships refer to the proximity of adjacent nodes in space, while dynamic correlations reflect similar traffic patterns exhibited by regions containing nodes with similar properties, such as schools, post offices and newspapers across different areas. Despite being geographically distant, these nodes exhibit similar peak traffic patterns.
	\begin{figure*}
		\centerline{\includegraphics[width=1\textwidth]{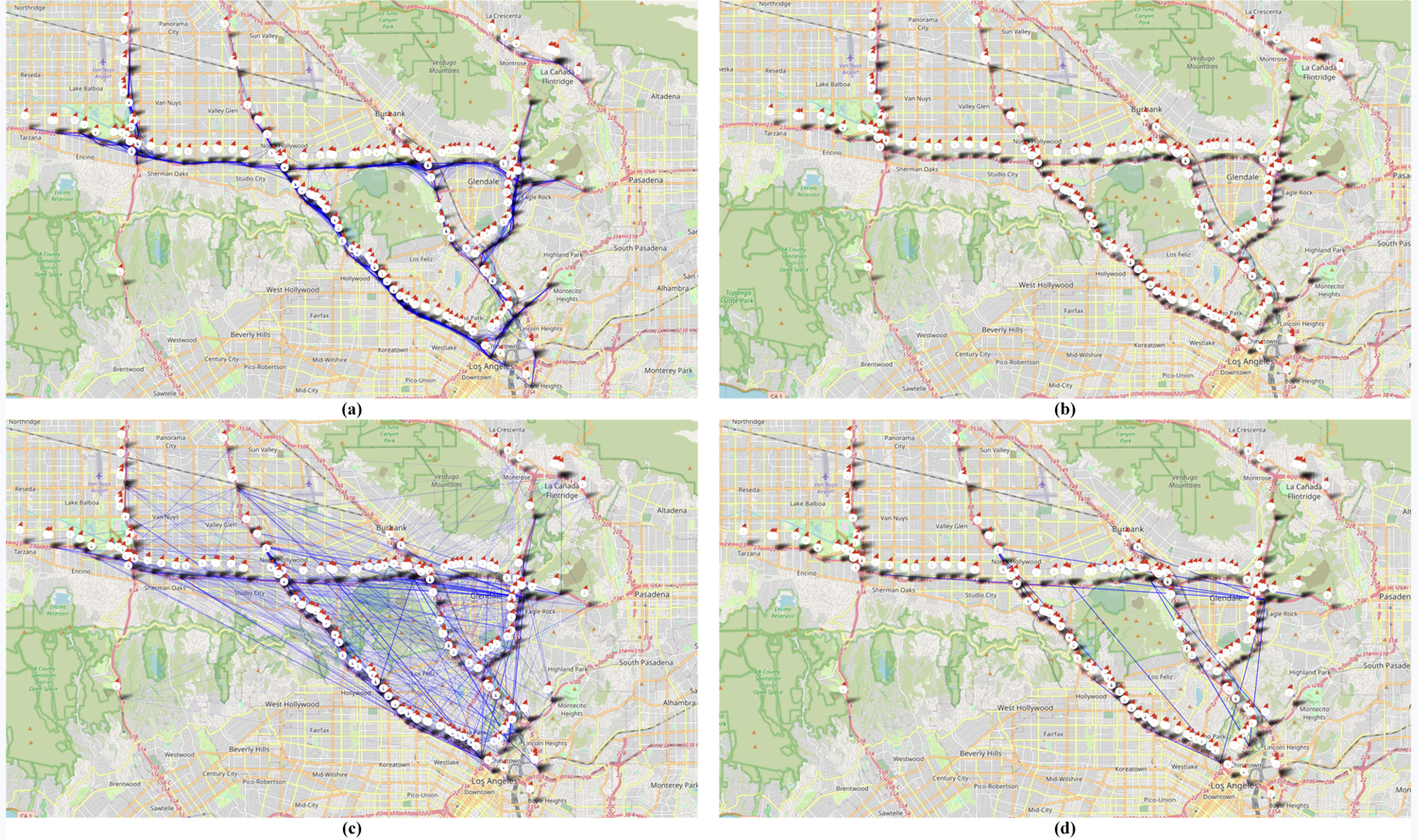}}
		\caption{Visualization of the road network in the METR-LA dataset.}
		\label{fig33}
	\end{figure*}
	
\begin{figure*}
		\centerline{\includegraphics[width=1\textwidth]{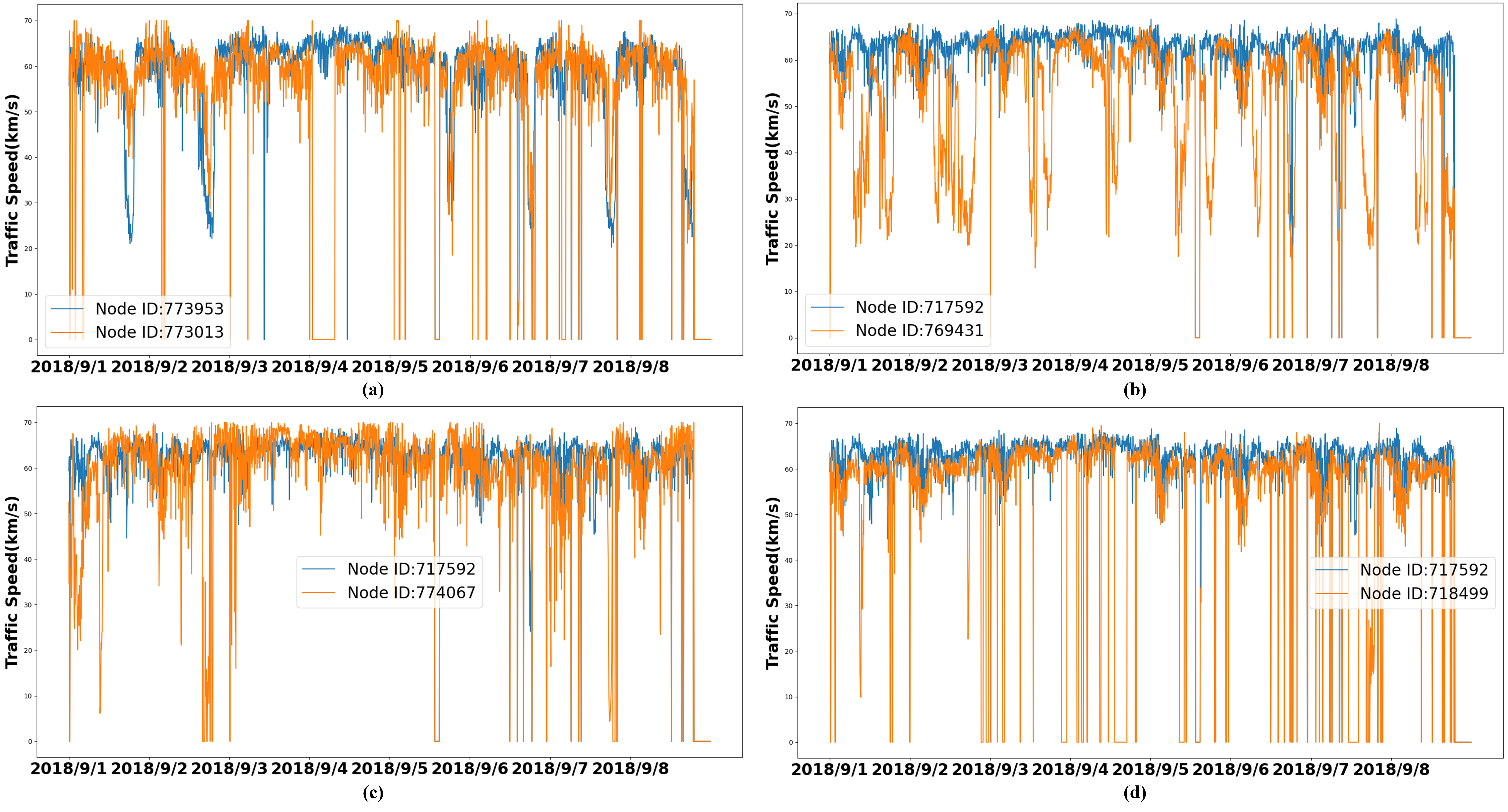}}
		\caption{Visualization of the speed curve of selected nodes in Fig. \ref{fig33}(d).}
		\label{fig44}
	\end{figure*}

	As shown in Fig. \ref{fig33}, in the METR-LA dataset, we visualize the structure of its road network. Fig. \ref{fig33}(a) The distribution of node positions and the correlation between nodes in the original road network can be seen. The spatial position correlations are reflected in the connected roads (nodes are connected by blue lines, and thicker lines indicate stronger correlation). Fig. \ref{fig33}(b) The inherent road network correlation learned in the adaptive adjacency matrix in this paper is relatively weak or even negligible, as the intrinsic structure is introduced. Therefore, the visualization results on the map are not obvious. Fig. \ref{fig33}(c) The unknown road network correlation learned in the adaptive adjacency matrix in this paper can be observed. Despite the long distance between nodes in the road network, some nodes still show strong correlation. Fig. \ref{fig33}(d) The few nodes with the strongest correlations after filtering. As shown in Fig. \ref{fig44}, it presents the speed curve diagrams of several selected nodes from Fig. \ref{fig33}(d). It can be observed that the trends between nodes are very similar, even though the connected nodes in Fig. \ref{fig33}(d) are far apart from each other. This further emphasizes the effectiveness of the adaptive adjacency matrix selected for feature learning in this paper.
	\section{Conclusion}
	In this paper, we propose a novel multi-channel spatial-temporal Transformer model for traffic flow forecasting. Our model adopts a multi-channel mechanism, which not only learns the static spatial features of the road network but also learns the dynamic spatial features, while introducing Transformers can better capture long-term dependencies. Experimental results on multiple real-world datasets demonstrate that MC-STTM outperforms state-of-the-art models, confirming the effectiveness of our approach in capturing spatial-temporal correlations. Specifically, the performance of deep learning models can be significantly improved by introducing a multi-channel mechanism. In the future, we will further explore how to capture more spatial-temporal information from raw data in cases of data missing or multi-modal data. On the other hand, with the development of multimodal data, future research will inevitably combine text information, image information, online information, and other sources to predict real-time traffic conditions.
	\section*{CRediT authorship contribution statement}
	The authors confirm contribution to the paper as follows: data collection and experiments: Baichao Long; theory and method: Jianli Xiao. All authors reviewed the results and approved the final version of the manuscript.
	\section*{Declaration of competing interes}
	The author(s) declared no potential conflicts of interest with respect to the research, authorship, and/or publication of this article.
	\section*{Data availability}
	The data used in this research is sourced from \cite{b26} and \cite{b27}, which is publicly available.
	\section*{Acknowledgments}
	This work is supported in part by China NSFC Program under Grant NO.61603257.

\end{document}